\documentclass[lettersize,journal]{IEEEtran}
\usepackage{amsmath,amssymb,amsfonts}
\usepackage{algorithmic}
\usepackage{algorithm}
\usepackage{array}
\usepackage[caption=false,font=normalsize,labelfont=sf,textfont=sf]{subfig}
\usepackage{textcomp}
\usepackage{stfloats}
\usepackage{url}
\usepackage{verbatim}
\usepackage{graphicx}
\usepackage{cite}
\usepackage{hyperref}
\hypersetup{hidelinks}
\usepackage{multirow}
\usepackage{textcomp}
\usepackage{cite}
\usepackage{bm}
\usepackage[normalem]{ulem}
\usepackage{threeparttable}
\useunder{\uline}{\ul}{}
\usepackage{booktabs}
\usepackage{romannum}

\begin{document}

\title{Higher-order Cross-structural Embedding Model for Time Series Analysis}

\author{Guancen Lin, Cong Shen and Aijing Lin
\thanks{This work has been supported by the National Natural Science Foundation of China (61673005) and China Scholarship Council (202307090068). (Corresponding author Cong Shen and Aijing Lin). }
\thanks{ Guancen Lin and Aijing Lin are with the School of Mathematics and Statistics, Beijing Jiaotong University, Beijing 100044, P. R. China. Guancen Lin is also with the School of Physical and Mathematical Sciences, Nanyang Technological University, Singapore, 637371.  (e-mail: 21118020@bjtu.edu.cn; ajlin@bjtu.edu.cn). }
\thanks{Cong Shen is with the Department of Mathematics, National University of Singapore, Singapore 119076, Singapore (e-mail: cshen@nus.edu.sg).}}

\markboth{Journal of \LaTeX\ Class Files,~Vol.~14, No.~8, August~2021}%
{Shell \MakeLowercase{\textit{et al.}}: A Sample Article Using IEEEtran.cls for IEEE Journals}


\maketitle

\begin{abstract}
Time series analysis has gained significant attention due to its critical applications in diverse fields such as healthcare, finance, and sensor networks. The complexity and non-stationarity of time series make it challenging to capture the interaction patterns across different timestamps. Current approaches struggle to model higher-order interactions within time series, and focus on learning temporal or spatial dependencies separately, which limits performance in downstream tasks. To address these gaps, we propose Higher-order Cross-structural Embedding Model for Time Series (High-TS), a novel framework that jointly models both temporal and spatial perspectives by combining multiscale Transformer with Topological Deep Learning (TDL). Meanwhile, High-TS utilizes contrastive learning to integrate these two structures for generating robust and discriminative representations. Extensive experiments show that High-TS outperforms state-of-the-art methods in various time series tasks and demonstrate the importance of higher-order cross-structural information in improving model performance.
\end{abstract}


\section{Introduction}
\label{Introduction}
Time series are pervasive across various domains, making the study of time series analysis critically important. Time series classification, in particular, has broad applicability in fields such as finance, healthcare and environmental monitoring \cite{gharehbaghi2017deep, gupta2020approaches, yang2022deep}. Accurate and efficient classification methods are essential for interpreting the internal patterns and making informed decisions based on the series. In recent times, deep learning has proven to be highly effective in time series classification, offering significant improvements in accuracy by automatically extracting and leveraging powerful representations from data \cite{liu2024diffusion, zhao2017lstm, shahid2022performance, wu2020comprehensive, scarselli2008graph}. 

\begin{figure}[] 
\centering
\includegraphics[width=0.45\textwidth]{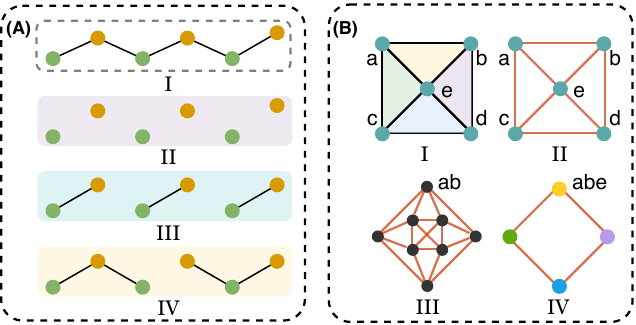} 
\caption{Higher-order patterns of temporal and spatial perspectives. (A) Higher-order patterns of temporal perspective. (B) Higher-order patterns of spatial perspective.}
\label{intro} 
\end{figure}

Time series inherently exhibit temporal and spatial dependencies. On the one hand, the temporal feature reflects the sequential nature of observations. In view of this characteristic, Recurrent Neural Network (RNN) and its variants such as LSTM and GRU are widely used \cite{medsker2001recurrent, schmidhuber1997long, cho2014learning}. Meanwhile, Transformer is introduced to leverage multi-head attention mechanisms, aiming to capture long-range dependencies in time series data \cite{vaswani2017attention, devlin2018bert, li2019enhancing}. On the other hand, spatial features represent the internal structure of the data, covering the complex interactions between different parts of the time series \cite{jin2024survey}. Mapping time series onto graphs is a common approach to capturing the spatial structure \cite{wang2024fully}. GNN-based methods can learn graphs mapped from time series to obtain effective representations. Consequently, a range of studies combine GNNs with temporal encoding to simultaneously capture the cross-structure of temporal and spatial dynamics \cite{jia2020graphsleepnet, wang2023sensor}. 

Cross-structural models that simultaneously consider the temporal and spatial domains have shown promising results in time series analysis \cite{wang2024fully, wang2024graph}. This inspires us to view sequential structures and graphs from different perspectives, exploring novel approaches to describing time series patterns. For example, subfigures \Romannum{1} of Fig. \ref{intro}(A) and Fig. \ref{intro}(B) respectively present a time series and its corresponding graph mapping. The traditional temporal perspective captures information from a single timestamp ((A)\Romannum{2}). Obviously, by aggregating multiple timestamps, new patterns emerge in (A)\Romannum{3} and (A)\Romannum{4}, which have not been considered in previous studies. Spatially, traditional methods capture the relationships between nodes ((B)\Romannum{2}). However, interactions beyond the node level are also deserve attention, such as constructing new structures through edges and triangles ((B)\Romannum{3} and \Romannum{4}). Recently, with the introduction of multiscale methods, higher-order patterns of multiple timestamps in the time domain have been captured, providing insights into short-term fluctuations and long-term trends. In the field of Topological Data Analysis (TDA), simplicial complexes take two or more vertices as a whole, extending the spatial properties of the graph to higher orders, thereby representing multiple levels of interactions.

In this paper, we introduce a novel representation learning framework named Higher-order Cross-structural Embedding Model for Time Series (High-TS), designed to integrate both temporal and spatial higher-order information in time series. Our High-TS model consists of two components: a multiscale embedding module for the temporal dimension and a simplicial complex embedding module for the spatial dimension. In the first module, a time-scale-based segmentation method is proposed, which adopts a multiscale attention mechanism to learn representations along the temporal dimension. In the second module, High-TS constructs higher-order interactions between different simplexes and uses Topological Deep Learning (TDL) to learn the new feature representation of each simplex. Finally, in view of the cross-structural perspectives of temporal and spatial dimensions, this paper employs contrastive learning to align representations of different modalities. This design facilitates the understanding of complex dynamics, thereby extracting features that reflect the intrinsic patterns of the data.

The main contributions of this paper are as follows:
\begin{itemize}
  \item Extending the temporal and spatial dimensions of time series from a single hierarchy to higher-order perspective through multiscale analysis and simplicial complexes.
  \item Combining Transformer and TDL to jointly learn short-term and long-term temporal dependencies across different scales, while capturing complex spatial interactions.
  \item Leveraging contrastive learning to integrate cross-temporal and spatial structures and enhance the representation of time series.
  \item Conducting extensive experiments to demonstrate the effectiveness of our approach to modeling and capturing complex interactions within time series.
\end{itemize}

\section{Related Works}
\subsection{Temporal Models for Time Series}
Temporal models have become integral to time series analysis, leveraging their ability to capture the temporal dependencies inherent in sequential data. RNNs were among the earliest deep learning models applied to time series data \cite{medsker2001recurrent}. However, RNNs suffer from issues like vanishing gradients, which limit their ability to model long-range dependencies \cite{bengio1994learning}. To address these challenges, LSTM networks \cite{schmidhuber1997long} and GRUs \cite{cho2014learning} were introduced. Recent variants such as Bidirectional LSTMs \cite{siami2019performance, wang2023dafa}, stacked LSTMs \cite{wang2020multi, essien2019deep} and Temporal Fusion Transformer (TFT) \cite{lim2021temporal, li2023probabilistic} have further improved performance in time series analysis tasks.

In addition to RNN-based approaches, 1D-CNNs have been adapted for time series analysis, which are particularly effective in capturing local patterns within the data \cite{zhao2017lstm, shahid2022performance}. More recently, attention-based models like Transformers \cite{vaswani2017attention}, and their variants such as BERT \cite{devlin2018bert}, Informer \cite{zhou2021informer} and TCN \cite{bai2018empirical} have gained popularity in time series analysis due to their ability to model complex dependencies without the need for sequential processing. In addition, there are some improved models based on self-attention mechanism, which have also achieved good results in time series analysis, such as ConvTrans \cite{mao2022convtrans}, ConvTrans-CL \cite{li2024convtrans}, Reformer \cite{kitaev2020reformer}, LogSparse Transformer \cite{li2019enhancing}, FEDformer \cite{zhou2022fedformer} and SCINet \cite{liu2022scinet}. Simultaneously, multiscale methods are increasingly recognized as an effective approach to time series analysis. Multiscale methods enable the examination of data across different temporal resolutions, revealing patterns and relationships that might be overlooked when considering a single scale, thereby enhancing model performance \cite{costa2002multiscale, li2023snake, li2024internal}. These advancements in deep learning architectures and time series analysis have significantly improved the accuracy and efficiency of time series prediction and classification, paving the way for more robust and scalable solutions in various domains.

\subsection{Graph and Simplicial Complex for Time Series}
The use of graph-based methods in time series analysis has emerged as a powerful approach to capturing complex dependencies. Different from traditional time series models, GNNs have been increasingly applied to time series analysis, leveraging the ability to model non-Euclidean structures \cite{wu2020comprehensive, scarselli2008graph}. Notable models include DCRNN\cite{li2018diffusion}, STGCN \cite{Yu2018Spatio}, Graph WaveNet  \cite{wu2019graph}, MTGNN \cite{wu2020connecting} and AGCRN \cite{bai2020adaptive}.

Beyond graphs, simplicial complexes extend the concept by modeling higher-order interactions among multiple nodes, which are not captured by pairwise relationships alone \cite{jonsson2008simplicial, popovic1997progressive}. Simplicial Neural Networks (SNNs) \cite{ebli2020simplicial} and topological methods have been introduced to exploit these higher-order structures in point cloud data. These methods allow for the analysis of interactions among more than two entities, offering a richer representation of the data \cite{schaub2020random}. Notable models include SCNN \cite{roddenberry2021principled}, which extends the idea of graph convolutional networks to simplicial complexes for modeling higher-order interactions. TDA techniques \cite{gidea2018topological} like persistent homology have also been used to analyze the topological features, providing insights that can inform more accurate analysis. Additionally, SAT \cite{goh2022simplicial} and PLL \cite{memoli2022persistent} use attention mechanisms, persistent Laplacian and simplicial complexes to focus on relevant higher-order interactions. Although these methods have made great progress in fields such as social networks, biomolecules, and materials science, their application in time series research is still rare. Thus, by integrating graph and simplicial complex-based methods, researchers are pushing the boundaries of time series analysis, enabling more accurate and insightful models that better capture the multifaceted nature of real-world data.

\section{Methodology}
\begin{figure*}[] 
\centering
\includegraphics[width=0.8\textwidth]{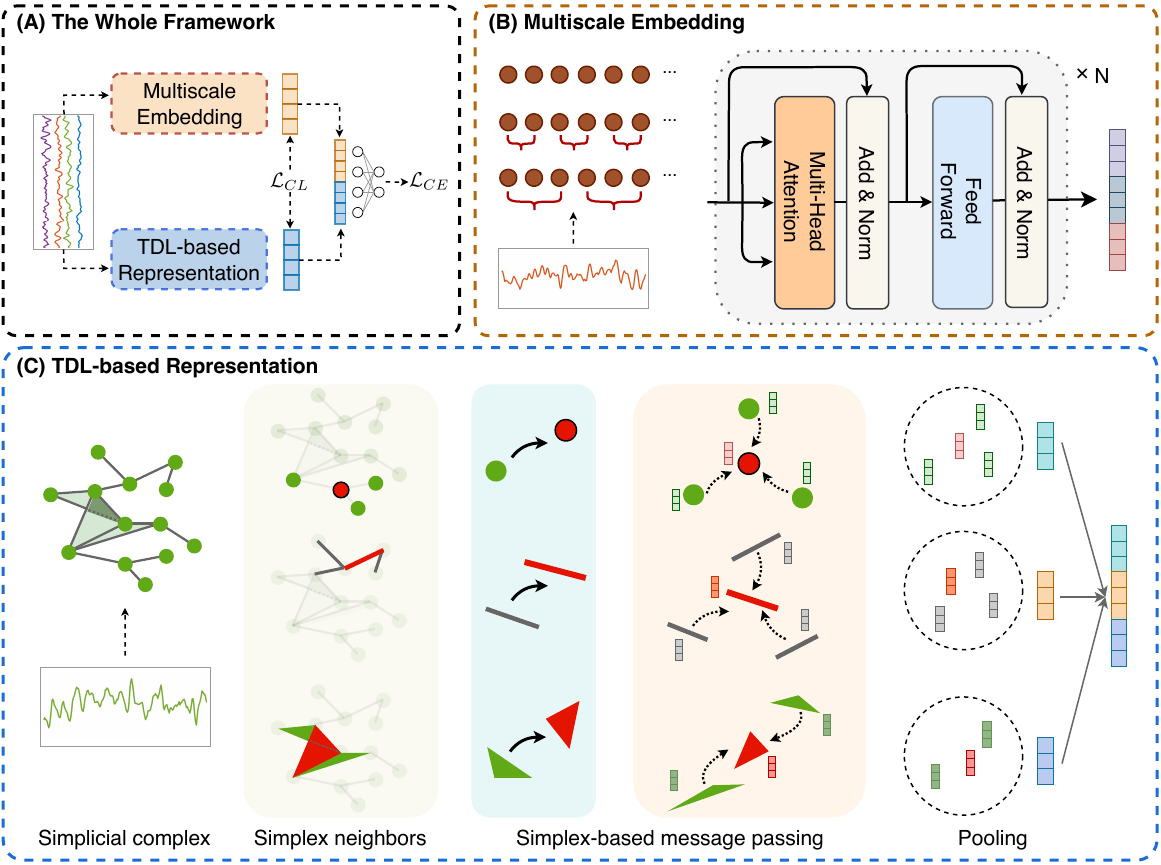} 
\caption{Overview of High-TS. (A) The whole framework of High-TS. Contrastive learning is constructed through multiscale embedding and TDL-based representation, followed by the output layer to obtain the final feature. (B) Multiscale embedding process. The combination of segmentation and encoder at multiple scales captures temporal embedding. (C) The TDL-based representation process. The passing of message between simplexes depicts the spatial structure.}
\label{overall} 
\end{figure*}
\subsection{Problem Formulation}
Suppose a time series dataset $\mathcal{D}$ containing $n$ labeled sequence samples  $\{\mathbf{x}_i, y_i\}_{i=1}^n$, where each sample  $\mathbf{x}_i \in \mathbb{R}^{L}$ is collected by a sensor across $T$ timestamps. Our task is to train an encoder $\mathcal{F}(\cdot)$ that is capable of simultaneously capturing the temporal and spatial interactions within the signals. Leveraging this encoder,  we can obtain a higher-order cross-structural embedding $\mathbf{r}_i=\mathcal{F}(\mathbf{x}_i) \in \mathbb{R}^d$ of the time series. By combining $\mathbf{r}_i$ with a classifier, we are able to perform the downstream task of classification, inferring the class $\hat{y}_i$ to which the sample $\mathbf{x}_i$ belongs. For simplicity,  $\mathbf{x}$ is employed instead of $\mathbf{x}_i$ in the following description, omitting the subscript $i$.

\subsection{Overall Structure}
Fig. \ref{overall} presents the overall architecture of High-TS, designed to capture higher-order cross-structural interactions within the time series. As shown in Fig. \ref{overall}(A), for each time series sample, we establish a dual-perspective encoding framework: the temporal perspective is derived from multiscale embeddings, while the spatial perspective is based on TDL. In the multiscale embedding module, the signal is first segmented into multiple segments according to pre-selected scales. These segments are processed by the Transformer encoder to capture higher-order attention. The features from different scales are concatenated to form the temporal embedding, as illustrated in Fig. \ref{overall}(B). Fig. \ref{overall}(C) outlines the TDL-based architecture. The time series is divided into multiple windows, each corresponding to a vertices in a simplex complex, with the number of vertices fixed. Message passing between neighboring higher-order simplexes is considered and the representations are concatenated to form a spatial embedding. Finally, Fig. \ref{overall}(A) illustrates how the temporal and spatial embeddings are aligned through contrastive learning and concatenated to produce the final time series representation for downstream tasks. More details are provided in subsequent sections.

\subsection{Multiscale Transformer Embedding}
Multiscale is a crucial approach in time series analysis, which examines data at different time scales to reveal hidden patterns and structures \cite{costa2002multiscale}. In time series analysis, Transformer effectively capture dependencies across different time stamps by utilizing the self-attention mechanism, making it well-suited for time series encoding \cite{shabani2023scaleformer, zhang2023crossformer}. In this section, we define an encoder $\mathcal{F_M}(\mathbf{x})$ for sample $\mathbf{x}$ that describes the integration of multiscale analysis with the Transformer for temporal representation of time series.

\subsubsection{Multiscale Sequence}
For a time series sample $\mathbf{x}=\{x_{(j)}\}_{j=1}^L \in \mathbb{R}^{L}$, we consider local temporal patterns by controlling a scale factor $s$ to segment the signal into multiple segments $\{\mathbf{h}_{t,s}\}_{t=1}^{L_s}$, where $L_s  = [\frac{L}{s}]$, $[\cdot]$ represents truncation operation. Each segment $\mathbf{h}_{t,s} =\{x_{(j)}\}_{j=(t-1) \times s+1}^{t \times s}$ of length $s$ is treated as an element of a multiscale time series, as shown in Fig. \ref{method}(A).

\subsubsection{Input Embedding and Positional Encoding}
Input embedding maps the raw time series segments into a higher-dimensional continuous latent space, enhancing the model's ability to capture subtle.
A linear layer is employed to learn the input embeddings for each segment of the multiscale time series. For a segment $\mathbf{h}_{t,s}$ at a specific scale $s$, the input embedding is represented as 
\begin{equation}
\mathbf{h}_{t,s}^I=\sigma(W_I \cdot \mathbf{h}_{t,s}+b_I),
\end{equation}
where $W_I$ is the weight matrix, and $b_I$ is the bias term. Additionally, to ensure that the temporal order of segments aligns with their corresponding positions in the original sequence, positional encoding $f_O(\cdot)$ is introduced as \cite{wang2024fully},
\begin{equation}
\begin{aligned}
\mathbf{h}_{t,s}^O=\{f_O^{(m)}(\mathbf{h}_{t,s})\}_{m=0}^{d}= 
\begin{cases}\sin (\omega_k \cdot t), \; m=2 k \\ 
\cos (\omega_k \cdot t), \; m=2 k+1,\end{cases}
\end{aligned}
\end{equation}
where $f_O^{(m)}(\mathbf{h}_{t,s})$ represents the positional encoding of the $m$-th position of $\mathbf{h}_{t,s}^I$, $d$ is the dimension of $\mathbf{h}_{t,s}^I$ and $\mathbf{h}_{t,s}^O$. Finally, the initial feature $\mathbf{u}_{t,s}$ of the segment $\mathbf{h}_{t,s}$ at scale $s$ is computed by combining the input embedding with the positional encoding, i.e., $\mathbf{u}_{t,s}=\mathbf{h}_{t,s}^I+\mathbf{h}_{t,s}^O$.

\subsubsection{Transformer Encoder-based Representation}
The encoder of Transformer  generates a continuous representation of the input sequence through a series of transformations based on attention mechanisms. Each encoder layer is divided into two parts, both of which are concluded with a residual connection and normalization (Add \& Norm) operation. In the first part, the queries $Q$, keys  $K$, and values  $V$ are applied to compute the attention score for each input according to Eq.\ref{QKV}.
\begin{equation}
A=\operatorname{Attention}(Q, K, V)=\operatorname{softmax}\left(\frac{Q K^T}{\sqrt{d_k}}\right) V.
\label{QKV}
\end{equation}
Multi-head attention allows combining representations from different spaces,
\begin{equation}
M=\operatorname{MultiHead}(Q, K, V) =\operatorname{Concat}( A_1, \ldots, A_h ) W^M,
\end{equation}
where $A_1, \ldots, A_h$ are the attention scores of the $h$-head obtained according to Eq.\ref{QKV}. For the initial feature $\mathbf{u}^s=\{\mathbf{u}_{t,s}\}_{t=1}^{L_s}$, it is encoded by the first part of the encoder as $N=\operatorname{LayerNorm}(M+\mathbf{u}^s)$. To preserve and optimize the information flow through the network, the second part combines a feed-forward neural network with the Add \& Norm operation. The raw time series is processed through the multiscale Transformer, as shown in Fig. \ref{overall}(B), resulting in its representation $\mathbf{z}^s$,
\begin{equation}
\mathbf{z}^s=\operatorname{LayerNorm}(\operatorname{FeedForward}(N)+N).
\end{equation}
And the temporal representation of sample $\mathbf{x}$ is defined as 
\begin{equation}
\mathcal{F_M}(\mathbf{x})=[\mathbf{z}^1 \oplus \cdots \oplus \mathbf{z}^s],
\end{equation}
where $\oplus$ means concatenation operation.

\begin{figure}[] 
\centering
\includegraphics[width=0.42\textwidth]{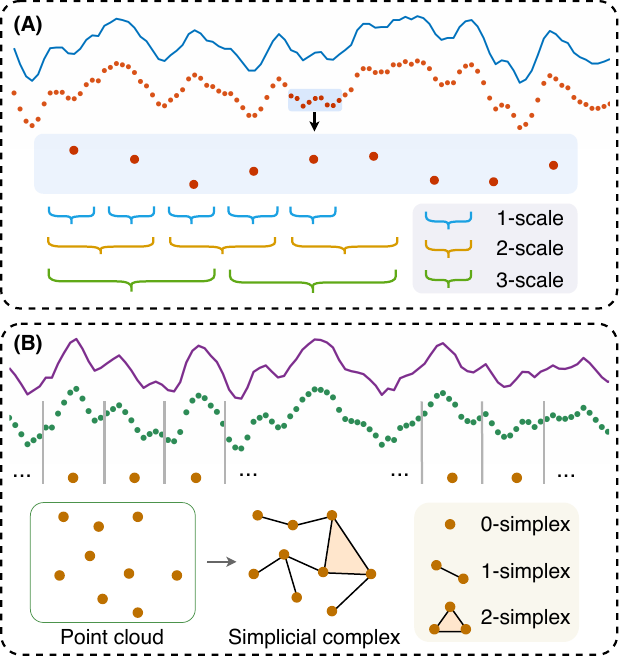} 
\caption{The formation process of multiscale sequence and simplicial complex. (A) The formation process of multiscale sequence. (B) The formation process of simplicial complex.}
\label{method} 
\end{figure}

\subsection{TDL-based Representation}
TDL \cite{hajijtopological, bodnar2023topological} leverages novel topological tools to characterize data with complicated higher-order structures. Different from graph-based data representation, TDL uses topological representations from algebraic topology, including cell complexes \cite{giusti2023cell}, simplicial complexes \cite{schaub2022signal}, and hypergraphs \cite{feng2019hypergraph}, to model not only pair-wise interactions (as in graphs), but also many-body or higher-order interactions among three or more elements. This section will focus on how to use simplicial complexes to represent time series and how to aggregate higher-order neighbor information. The TDL-based encoder is introduced as $\mathcal{F_T}(\mathbf{x})$.

For a time series $\mathbf{x}=\{x_{(j)}\}_{j=1}^L \in \mathbb{R}^{L}$, a point number $n$ is introduced to constrain the patch size $L_p$, where the time series is segmented such that $L_p=[\frac{L}{n}]$. Each patch $\mathbf{p}_{t} =\{x_{(j)}\}_{j=(t-1) \times L_p+1}^{t \times L_p}$ of length $L_p$ is introduced as a point of the point cloud $P=\{\mathbf{p}_t\}_{t=1}^n \in \mathbb{R}^{n}$.

\subsubsection{Simplicial Complex for Time Series}
A simplicial complex is the generalization of a graph into its higher-dimensional counterpart. The simplicial complex is composed of simplexes. Each simplex is a finite set of vertices and can be viewed geometrically as a point (0-simplex), an edge (1-simplex), a triangle (2-simplex), a tetrahedron (3-simplex), and their k-dimensional counterpart ($k$-simplex), as shown in Fig. \ref{method}(B). More specifically, a $k$-simplex $\sigma^k=\{v_0,v_1,v_2,\cdots, v_k\}$ is the convex hull formed by $k+1$ affinely independent points $v_0,v_1,v_2,\cdots, v_k$ as follows
\begin{equation} 
\sigma^{k} = \left\{ \lambda_{0}v_{0} + \lambda_{1}v_{1} + \cdots + \lambda_{k}v_{k} \middle| {\sum\limits_{i = 0}^{k}{\lambda_{i} = 1;\forall i,0 \leq \lambda_{i} \leq 1}} \right\}.
\end{equation}
The $i$-th dimensional face of $\sigma^{k}(i<k)$ is the convex hull formed by $i+1$ vertices from the set of $k+1$ points $v_0,v_1,v_2,\cdots, v_k$. The simplexes are the basic components for a simplicial complex.

A simplicial complex $K$ is a finite set of simplexes that satisfy two conditions. First, any face of a simplex from $K$ is also in $K$. Second, the intersection of any two simplexes in $K$ is either empty or a shared face. A $k$-th chain group $C_k$ is an Abelian group of oriented $k$-simplexes $\sigma^{k}$, which are simplexes together with an orientation, i.e., ordering of their vertex set. The boundary operator $\partial_{k}( C_{k}\rightarrow C_{k-1} )$ for an oriented $k$-simplex $\partial_{k}$ can be denoted as
\begin{equation}
\partial_{k}\sigma^{k} = {\sum\limits_{i = 0}^{k}{(-1)^{i}\left\lbrack v_{0},v_{1},v_{2},\cdots,{\hat{v}}_{i},\cdots,v_{k} \right\rbrack}}.
\end{equation}
Here, $\left\lbrack v_{0},v_{1},v_{2},\cdots,{\hat{v}}_{i},\cdots,v_{k} \right\rbrack$ is an oriented $(k-1)$-simplex, which is generated by the original set of vertices except $v_i$. The boundary operator maps a simplex to its faces, and it guarantees that $\partial_{k-1}\partial_{k} = 0$.

In this study, we employ the Vietoris-Rips complex to describe both the topological and geometric structure of a given point cloud. Formally, the Vietoris-Rips complex, denoted as $\mathrm{Rip}=\mathrm{Rip}(P)$, is defined for a point cloud $P$ and a cutoff value $c$, as follows
\begin{equation} 
\begin{aligned}
\mathrm{Rip}(P) := \{ \sigma \subseteq P \mid \sigma \text{ is finite} & \text{ and } \forall \mathbf{p}_i, \mathbf{p}_j \in \sigma, \;  \\
& \operatorname{sim}(\mathbf{p}_i, \mathbf{p}_j) \geq c \},
\end{aligned}
\end{equation}
where $\sigma$ represents a simplex in $P$, and $\operatorname{sim}(\mathbf{p}_i, \mathbf{p}_j)$ denotes the cosine similarity between $\mathbf{p}_i$ and $\mathbf{p}_j$.

In this section, we consider these relations as interactions between $k$-simplexes. Specifically, we focus on the upper adjacency relation for all $0$-simplexes. This means an interaction occurs between two points if and only if their similarity is no less than $c$. When considering higher-order $k$-simplexes (where $k>0$), we focus on their lower adjacency; that is, two $k$-simplexes interact if and only if their common face is a $(k-1)$-simplex. We use adjacency matrix $A_{k}=\{A_{c,k}(i,j)\}$ to describe this interaction between $k$-simplexes in $\mathrm{Rip}$:
\begin{equation}
A_{k}(i,j)=\left\{ \begin{matrix}
{1,~~~\sigma_{i}^{0} \smallfrown \sigma_{j}^{0},i \neq j,k = 0~~~} \\
{1,~~~\sigma_{i}^{k} \smallsmile \sigma_{j}^{k},i \neq j,k \geq 1~~~} \\
{0,~~~others.~~~~~~~~~~~~~~~~~~~}
\end{matrix} \right. 
\end{equation}
where $\sigma_{i}^{k}$ is the $i$-th $k$-simplex in $\mathrm{Rip}$. And the notation $\smallfrown$ and $\smallsmile$ represent the upper and lower adjacency relation, respectively.

Also we consider the following diagonal matrices $D_k=\{D_{c,k}(i,j)\}$ for normalization:
\begin{equation}
D_{k}(i,j)=\left\{ \begin{matrix}
{d_U(\sigma_{i}^{0}),~~i = j,~~k = 0} \\
{d_L(\sigma_{i}^{k}),~~i = j,~~k \geq 1} \\
{0~~~~~~~ , ~~others.~~~~~~~}
\end{matrix} \right. 
\end{equation}
where $d_U(\sigma_{i}^{k})$ and $d_L(\sigma_{i}^{k})$ denote the upper and lower degree of $\sigma_{i}^{k}$, respectively.

\subsubsection{Simplex-based Message Passing}
In High-TS, Vietoris-Rips complexes $\mathrm{Rip}$ is utilized to describe time series. To be specific, we consider a series of adjacent matrices $A_{k}$ to describe the interaction between $k$-simplexes in $\mathrm{Rip}$. We use message passing to learn the feature representation of each simplex,
\begin{equation} 
H_{k}^{(l + 1)} = {\rm Relu}\left({\hat{D}}_{k}^{- \frac{1}{2}}{\hat{A}}_{k}{\hat{D}}_{k}^{-\frac{1}{2}}H_{k}^{(l)}W_{k}^{(l)} \right),
\end{equation}
and the initial representations $H_0^{(0)}$, $H_1^{(0)}$ and $H_2^{(0)}$ of 0-samplex, 1-samplex, and 2-samplex are shown in Section A of the supplementary materials.

In the $l$-th iteration, the feature matrix $H_{k}^{(l + 1)}$ of $k$-simplexes is obtained by gathering neighbors feature of each $k$-simplex. Here $\hat{A}_{k}$ represents the sum of $A_{k}$ and identity matrix. $\hat{D}_{k}$ is a degree matrix, which is a diagonal matrix whose values on the diagonal are equal to the sum of the corresponding rows (or columns) in $\hat{A}_{k}$. $W_{k}^{(l)}$ is the weight matrix (to be learned). Computationally, we usually repeat the process 1 to 3 times, and the final simplex feature is denoted as $H_{k} $.
After message passing, all $k$-simplex features in $\mathrm{Rip}$ are gathered into one feature through a pooling process,
\begin{equation}
\mathbf{f}_{k} ={\rm Pooling} \left(H_{k}\right),
\end{equation}
where ${\rm Pooling}(\cdot)$ is a pooling function. And the spatial representation of sample $\mathbf{x}$ is defined as 
\begin{equation}
\mathcal{F_T}(\mathbf{x})=[\mathbf{f}^0 \oplus \cdots \oplus \mathbf{f}^k].
\end{equation}
The whole process of the simplex-based message massing is shown in Fig. \ref{overall}(C).

\subsection{Cross-structural Contrastive Learning}
\label{CL}
Contrastive learning is effectively applied in unsupervised settings by utilizing different views or augmentations of the same sample, enabling the model to learn meaningful representations without the need for labeled data \cite{you2020graph}. After obtaining the embeddings from the multiscale Transformer and TDL encoding, we employ contrastive learning to enhance the ability of our framework to learn more discriminative representations. 

In this paper, we combine the unsupervised aspect of contrastive learning with the cross-structural patterns in High-TS, training the network to minimize the distance between positive pairs and maximize the distance between negative pairs \cite{you2020graph}. A batch containing $B$ time series samples is randomly selected, yielding $2B$ augmented representations through cross-structural encoders $\mathcal{F_M}(\cdot)$ and $\mathcal{F_T}(\cdot)$. Specifically, we obtain augmented representations of samples through a cross-structural mechanism. For a sample $\mathbf{x}_i$ in the batch, its embedding $\mathcal{F_M}(\mathbf{x}_i)$ from the multiscale Transformer and the embedding $\mathcal{F_T}(\mathbf{x}_i)$ from TDL form a positive pair, while the augmented representations of different samples in the batch are treated as negative pairs. The contrastive loss of $\mathbf{x}_i$ is defined as follows:
\begin{equation}
\begin{aligned}
& \ell_{CL}(\mathbf{x}_i)=-\left(\log \frac{\exp (\operatorname{sim}(\mathcal{F_M}(\mathbf{x}_i), \mathcal{F_T}(\mathbf{x}_i)) / \tau)}{\sum_{j=1, j \neq i}^B \exp (\operatorname{sim}(\mathcal{F_M}(\mathbf{x}_i), \mathcal{F_T}(\mathbf{x}_j) / \tau)} \right. \\
& \left. +\log \frac{\exp (\operatorname{sim}(\mathcal{F_M}(\mathbf{x}_i), \mathcal{F_T}(\mathbf{x}_i)) / \tau)}{\sum_{j=1, j \neq i}^B \exp (\operatorname{sim}(\mathcal{F_M}(\mathbf{x}_j), \mathcal{F_T}(\mathbf{x}_i) / \tau)}\right).
\end{aligned}
\end{equation}
where $\tau$ denotes the temperature parameter, $\operatorname{sim(\cdot,\cdot)}$ is used to measure the cosine similarity between two representations. The final loss is computed across all positive pairs within the batch,
\begin{equation}
\mathcal{L}_{CL}=\frac{1}{2} \sum_{i=1}^B(\ell_{CL}(\mathbf{x}_i)).
\end{equation}

\subsection{Fusion and Prediction}
In addition to using unsupervised contrastive learning for time series representation learning, we also designed a supervised time series classification task in this section. After the cross-structural encoding process, the higher-order representations from the temporal and spatial dimensions, $\mathcal{F_M}(\mathbf{x})$ and $\mathcal{F_T}(\mathbf{x})$, are used to represent the sample features from two different perspectives. These representations are first concatenated and then fed into a fully connected layer for label prediction. The connected representation of the sample is denoted by $\mathbf{r}$, and $\hat{y}=\{\hat{y_i}\}_{i=1}^n$ represents the predicted label obtained by the fully connected layer.
\begin{equation}
\begin{aligned}
\mathbf{r} = \mathcal{F}(\mathbf{x}) & =\left[\mathcal{F_M}(\mathbf{x}) \oplus \mathcal{F_T}(\mathbf{x}) \right] \\
& =\left[\mathbf{z}^1 \oplus \mathbf{z}^2 \oplus \mathbf{z}^3 \oplus \mathbf{f}_0 \oplus \mathbf{f}_1 \oplus \mathbf{f}_2 \right],
\end{aligned}
\end{equation}
\begin{equation}
\hat{y}=\sigma(W_R \cdot \mathbf{r}+b_R).
\end{equation}

We designate cross-entropy loss as the objective function to optimize the parameters within the framework,
\begin{equation}
\mathcal{L}_{CE}=-\sum_{i=1}^B y_i \log (\hat{y}_i)+(1-y_i) \log (1-\hat{y}_i).
\end{equation}
Together with the loss of contrastive learning in Section \ref{CL}, the overall loss function $\mathcal{L}$ in this paper is defined as:
\begin{equation}
\mathcal{L}=\mathcal{L}_{CE}+\mathcal{L}_{CL}.
\end{equation}
Guided by labeled data, we are able to refine the proposed model through back propagation, enabling the learning of sample embeddings for classification.

\section{Experimental Evaluation}
\label{Experimental}
\subsection{Datasets and Baselines}
To validate the effectiveness of the High-TS framework, we select 12 publicly available time series datasets from various real-world fields. These datasets encompass a wide range of applications, providing a comprehensive evaluation of our model's performance across different types of time series data. The details of the datasets are shown in Table \ref{dataset}. We conducted extensive experiments, comparing the results with state-of-the-art (SOTA) methods to demonstrate the superiority of our approach to capturing complex temporal and spatial interactions.
\subsubsection{UCR Time Series Archive}
We utilized 10 datasets from the UCR dataset\footnote{https://timeseriesclassification.com/dataset.php}, namely DistalPhalanxOutlineAgeGroup (DPOAG), DistalPhalanxOutlineCorrect (DPOC), ECG5000, FreezerRegularTrain (FRT), Ham, MiddlePhalanxOutlineCorrect (PPOC), ProximalPhalanxOutlineAgeGroup (PPOAG), RefrigerationDevices (RD), Strawberry, and Wine.
\subsubsection{Epilepsy Electroencephalogram Archive}
The Epileptic electroencephalogram (EEG) dataset \cite{andrzejak2001indications} records EEG signal segments from 500 subjects, with each segment lasting 23.6 seconds and a sampling frequency of 173.61 Hz. These segments are categorized into five groups: A and B represent healthy subjects with eyes open and closed, respectively; C and D are recordings from the hippocampal structure in the opposite hemisphere and the epileptogenic zone during the interictal phase of epilepsy patients; E represents the ictal period. In this study, we create a diagnostic task for epilepsy, aiming to distinguish between interictal (C) and ictal (E) periods, as well as between healthy states (B), interictal (C), and ictal (E) periods. The preprocessing of the raw data followes reference \cite{eldele2021time}.
\subsubsection{Baseline Methods}
We conduct experiments on these datasets to compare the classification performance of our model against SOTA algorithms, which cover the fields of deep learning, machine learning, and statistics: GGD-RCMDE (G-R) combined with SVM \cite{dhandapani2022novel}, HDE combined with RF \cite{wang2022hierarchical}, DTW \cite{chen2013dtw}, TNC \cite{tonekaboni2021unsupervised}, TST \cite{zerveas2021transformer}, TS-TCC \cite{eldele2021time}, T-Loss \cite{franceschi2019unsupervised}, TS2Vec \cite{yue2022ts2vec}, ROCKET \cite{dempster2020rocket} and ShapeConv \cite{qu2024cnn}.

\begin{table}[]
\caption{Description of datasets used in the experiments. Including the number of training samples (\# Train) and testing samples (\# Test), series length, and the number of classes (\# Class) for each dataset.}
\centering
\setlength{\tabcolsep}{8pt}
\renewcommand\arraystretch{1.2} 
\begin{tabular}{l|l|cccc}
\toprule
Source                    & Datasets   & \# Train & \# Test & Length & \# Class \\
 \midrule
\multirow{8}{*}{UCR}      & DPOAG      & 400      & 139     & 80     & 3        \\
                          & DPOC       & 600      & 276     & 80     & 2        \\
                          & ECG5000       & 500      & 4500     & 140     & 5        \\
                          & FRT         & 150      & 2850   & 301    & 2        \\
                          & Ham        & 109      & 105     & 431    & 2        \\
                          & MPOC       & 600      & 291     & 80     & 2        \\
                          & PPOAG      & 400      & 205     & 80     & 3        \\
                          & RD         & 375      & 375     & 720    & 3        \\
                          & Strawberry & 613      & 370     & 235    & 2        \\
                          & Wine       & 57       & 54      & 234    & 2        \\
\midrule
\multirow{2}{*}{Epilepsy} & C\&E       & 3680     & 920     & 178    & 2        \\
                          & B\&C\&E     & 5520     & 1380    & 178    & 3       \\
\bottomrule 
\end{tabular}
\label{dataset}
\end{table}

\subsection{Experimental Setup}
We partitioned 80\% of the Epilepsy Electroencephalogram archive as the training set and 20\% as the test set, with 25\% of the training data reserved for validation to choose appropriate hyperparameters. Accuracy is used as the evaluation metric for model performance. The experiments are repeated 5 times using different random seeds, and the mean accuracy along with the standard deviation is reported. The experiments utilize the Adam optimizer with a batch size of 64, a learning rate of 2e-4, and 3e+3 epochs. The cutoff $c$ is set as the cosine similarity value located in the top 10\% of all cosine similarities. We employ grid search to select the optimal combination of the number of vertices and the dimension of the latent representation, where the number of vertices is chosen from $\{15, 20, 25, 30\}$, and the latent dimension is evaluated over $\{8, 16, 32, 64\}$. Experiments are conducted on a computational setup featuring a 32-core AMD 75F3 CPU, 500GB RAM, and NVIDIA A100 40G GPU capabilities. The server infrastructure is provided by the National Supercomputing Centre (NSCC) Singapore.

\begin{table*}[]
\caption{High-TS compared with SOTA time-series classification methods. The best result in each row is highlighted in bold, while the second-best result is indicated with an underline. Each value in parentheses represents the standard deviation from five experimental evaluations.}
\centering
\setlength{\tabcolsep}{2pt}
\renewcommand\arraystretch{1.4} 
\begin{threeparttable}
\begin{tabular}{p{1.5cm} | p{1.1cm}<{\centering} p{1.1cm}<{\centering} p{1.1cm}<{\centering} p{1.1cm}<{\centering} p{1.1cm}<{\centering} p{1.2cm}<{\centering} p{1.1cm}<{\centering} p{1.1cm}<{\centering} c p{1.7cm}<{\centering} p{1.7cm}<{\centering}}
\toprule
\multirow{2}{*}{Datasets} & \multicolumn{11}{c}{Methods}                                                                                                                     \\ \cline{2-12} 
                          & G-R & HDE   & DTW   & TNC         & TST   & TS-TCC & T-Loss      & TS2Vec      & ROCKET      & ShapeConv            & High-TS                  \\ 
\midrule
DPOAG                     & 0.698     & 0.518 & 0.770 & 0.741       & 0.741 & 0.755  & 0.727       & 0.727       & 0.759       & {\ul $0.784_{(0.017)}$}    & \pmb{$0.804_{(0.032)}$} \\
DPOC                      & 0.634     & 0.646 & 0.717 & 0.754       & 0.728 & 0.754  & {\ul 0.775} & {\ul 0.775} & 0.770       & $0.753_{(0.026)}$          & \pmb{$0.807_{(0.025)}$} \\
ECG5000                     &  0.862     & 0.691     & 0.924 & 0.937      & 0.928 & 0.941 & 0.933 & 0.935 & {\ul 0.947}   & \pmb{$0.953_{(0.007)}$}    & $0.942_{(0.002)}$ \\
FRT                        & 0.772      &  0.693     & 0.899 & 0.991 & 0.922 & 0.989 & 0.956 & 0.986 & \pmb{0.998} & {\ul $0.993_{(0.003)}$} & \pmb{$0.998_{(0.000)}$}           \\
Ham                       & 0.629     & 0.669 & 0.467 & {\ul 0.752} & 0.524 & 0.743  & 0.724       & 0.724       & 0.726       & $0.733_{(0.032)}$          & \pmb{$0.798_{(0.049)}$} \\
MPOC                      & 0.574     & 0.568 & 0.698 & 0.818       & 0.753 & 0.818  & 0.825       & {\ul 0.838} & {\ul 0.838} & $0.827_{(0.004)}$          & \pmb{$0.847_{(0.014)}$} \\
PPOAG                     & 0.854     & 0.563 & 0.805 & 0.854       & 0.854 & 0.839  & 0.844       & 0.844       & 0.856       & {\ul $0.869_{(0.004)}$}    & \pmb{$0.878_{(0.006)}$} \\
RD                        & 0.587     & 0.447 & 0.464 & 0.565       & 0.483 & 0.563  & 0.515       & 0.589       & 0.537       & {\ul $0.594_{(0.026)}$}    & \pmb{$0.624_{(0.031)}$} \\
Strawberry                & 0.719     & 0.752 & 0.941 & 0.951       & 0.916 & 0.965  & 0.954       & 0.965       & {\ul 0.981} & $0.903_{(0.008)}$          & \pmb{$0.984_{(0.004)}$} \\
Wine                      & 0.500     & 0.596 & 0.574 & 0.759       & 0.500 & 0.778  & 0.815       & 0.889       & 0.813       & {\ul $0.894_{(0.018)}$}    & \pmb{$0.974_{(0.017)}$} \\ 
C\&E                      & 0.729     & 0.604 & —   &    0.626    & 0.966 & 0.917  & 0.969       & {\ul 0.985}       & 0.911       & $0.877_{(0.010)}$           & \pmb{$0.994_{(0.001)}$}            \\
B\&C\&E                    & 0.764     & 0.421 & —     &    0.420    & 0.770 & 0.943  &   0.926     & {\ul 0.949}       & 0.723       & $0.829_{(0.019)}$          & \pmb{$0.973_{(0.004)}$}            \\ 
\midrule
Average                   & 0.693       &  0.597       & 0.726      &  0.764    & 0.757  	 & 0.834 	& 0.830 	&  {\ul 0.851} 	& 0.822 	& 0.834       &    \pmb{0.885}                  \\ 
\bottomrule
\end{tabular}
\label{results}
\begin{tablenotes} 
\footnotesize  
\item[—] The performance on C\&E and B\&C\&E is not reported due to the unavailability of the code for DWT.
\end{tablenotes}        
\end{threeparttable}  
\end{table*}

\begin{figure*}[] 
\centering
\includegraphics[width=1\textwidth]{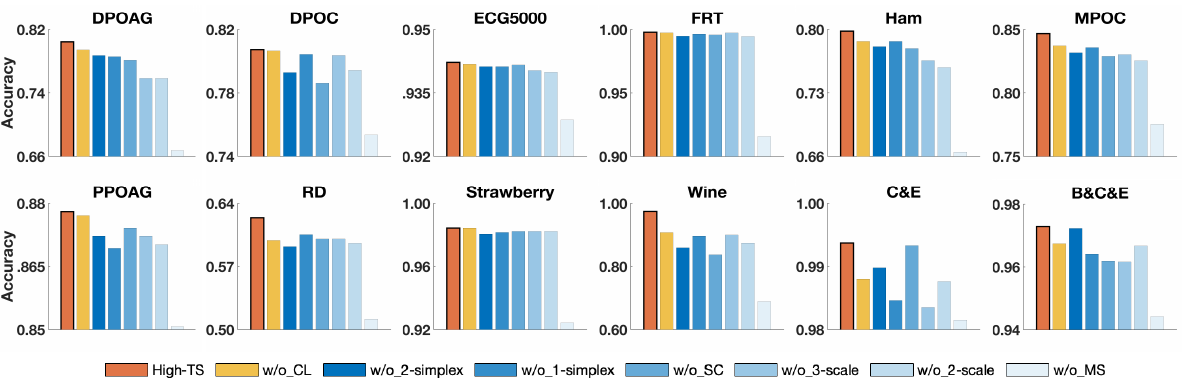} 
\caption{Ablation study of the effect of different components in High-TS on 12 datasets.}
\label{ablation} 
\end{figure*}

\subsection{Results and Analysis}
To evaluate the performance of the proposed framework, we execute High-TS and SOTA methods across all datasets, and the experimental results are presented in Table \ref{results}. The classification accuracy demonstrates that High-TS outperforms all baseline methods on 11 out of the 12 datasets, achieving the best performance. In particular, the average classification accuracy for each model is provided, and High-TS demonstrates a clear advantage over the second-place method by a margin of 4.09\%, highlighting the strengths of our framework. Specifically, our model shows the greatest advantage on the Wine dataset, followed by the Ham and RD datasets, with accuracy improvements of 8.95\%, 6.12\%, and 5.05\% over the second-best baseline, respectively. Furthermore, High-TS is not only effective for datasets with abundant samples but also demonstrates advantages on datasets with few samples, and achieves high classification accuracy for both long and short time series. Although High-TS falls short of the top method on the ECG5000 dataset, it still ranks among the top three, with accuracy only 1.15\% below the highest value. This slight underperformance may be attributed to the significantly smaller training set size compared to the test set for this dataset. Similarly, the result on the FRT dataset is only on par with the SOTA method. Overall, High-TS exhibits robust performance across various types of time series data, surpassing advanced statistical, machine learning, and deep learning algorithms in classification accuracy. These findings underscore the necessity of capturing higher-order cross-structural interactions and validate the effectiveness of the High-TS framework. (The accuracy of the third to tenth baseline methods in Table \ref{results} on the UCR datasets is sourced from reference \cite{qu2024cnn}.)

\subsection{Ablation study}
We evaluate the effectiveness of each component in High-TS through ablation studies. First, we train the model without contrastive learning, referred to as 'w/o\_CL'. On this basis, we focus on the role of higher-order, separately remove the 2-simplex and 1-simplex components, denoted as 'w/o\_2-simplex' and 'w/o\_1-simplex'. Additionally, we omit the designed 3-scale and 2-scale components to create the variants 'w/o\_3-scale' and 'w/o\_2-scale', respectively. Finally, to validate the importance of cross-structural integration, we train variants 'w/o\_SC' with only multiscale Transformer without TDL and variant 'w/o\_MS' with only TDL without multiscale Transformer. These variants are compared with the complete version, and the results are reported in Fig. \ref{ablation}.

Evidently, contrastive learning contributes to generating robust features for the classification task. For instance, in the Wine and RD datasets, the complete version achieves improvements of 7.35\% and 4.19\%, respectively, over the 'w/o\_CL' variant. This highlights the necessity of the cross-structural modules acting as mutual augmentations. At the same time, by reducing higher-order information, the performance of the 'w/o\_2-simplex', 'w/o\_1-simplex', 'w/o\_3-scale', and 'w/o\_2-scale' variants shows a decline compared to 'w/o\_CL'. Additionally, with the introduction of cross-structural components, the performance of 'w/o\_2-simplex' and 'w/o\_1-simplex' notably surpasses that of the 'w/o\_SC', which utilizes only temporal patterns, with classification accuracy increasing by 0.83\% and 2.30\% on the DPOC dataset. Similarly, 'w/o\_3-scale' and 'w/o\_2-scale' further enhanced the performance of the 'w/o\_MS' variant, which operates solely from the spatial perspective, showing significant improvements on the DPOAG, Ham, and RD datasets. Notably, 'w/o\_MS' consistently demonstrates the weakest performance across all experiments, indicating that spatial information alone is insufficient to fully capture the characteristics of time series.

These observations are also consistent across other datasets, further highlighting the importance of higher-order cross-structural patterns and contrastive learning. This combination significantly contributes to the overall superior performance in downstream tasks. Detailed results are given in Section B.1 of the supplementary materials.

\begin{figure}[h] 
\centering
\includegraphics[width=0.48\textwidth]{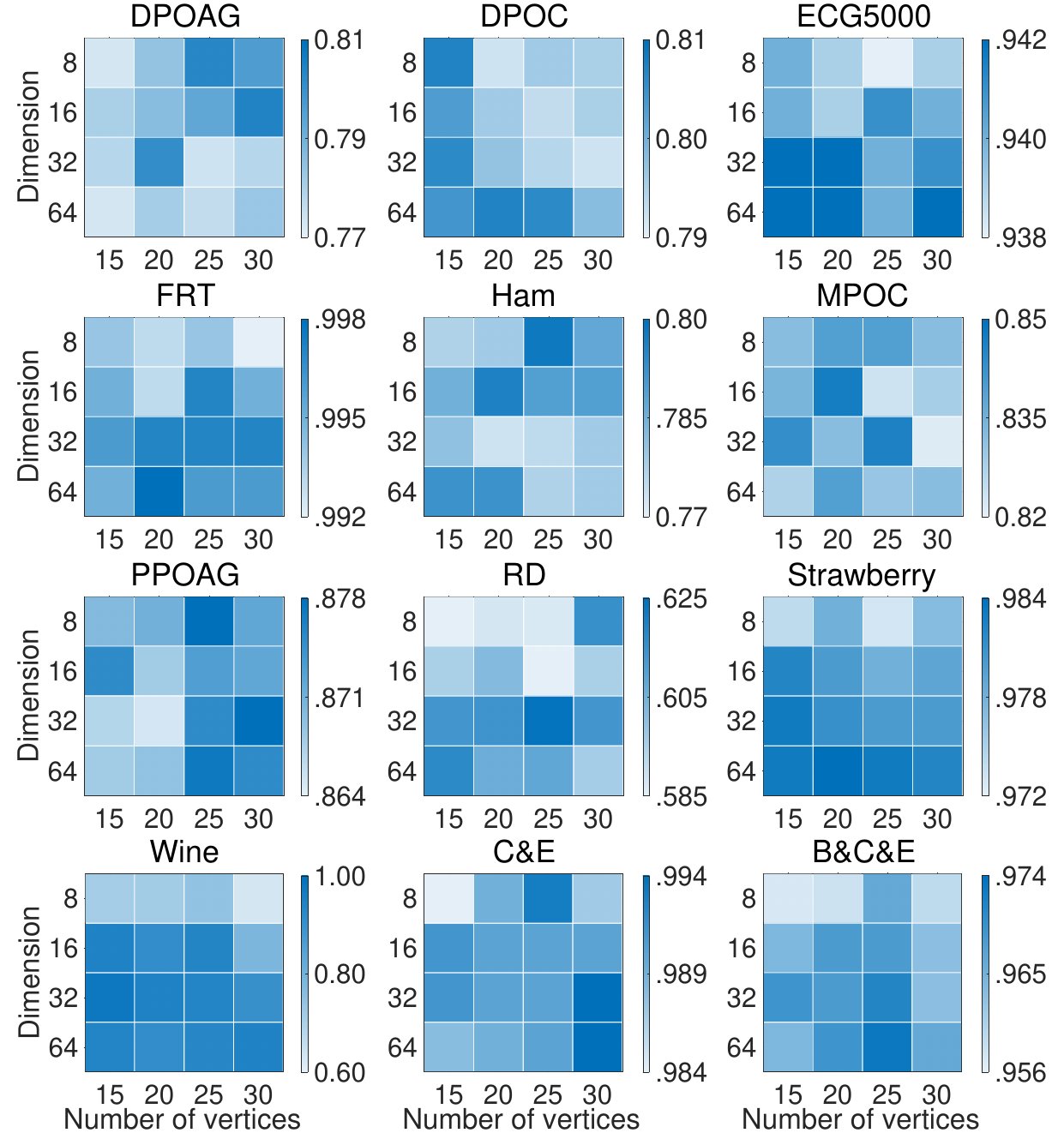} 
\caption{Results of grid search on the number of vertices and the dimension of latent representation on 12 datasets.}
\label{grid} 
\end{figure}

\subsection{Parameters Sensitivity Study}
In this section, we discuss the impact of latent representation dimensions and the number of vertices in High-TS, with the results shown in Fig. \ref{grid}. For the latent representation dimension, we observe that datasets with larger sample sizes tend to achieve better performance with higher dimensionality. Specifically, the largest datasets, B\&C\&E and C\&E from the Epilepsy Electroencephalogram archive, show significantly higher accuracy at 64 and 32 dimensions compared to dimension 8. This trend is also evident in the Strawberry dataset, which has the third-largest training set size. These results suggest that larger latent dimensions can capture more information across samples.

For the number of vertices, the results show that the accuracy generally improves as the number of vertices increases, however, the performance decreases at the highest vertex count, as seen in datasets such as FRT, Ham, PPOAG, RD, C\&E, and B\&C\&E. This phenomenon may be due to the fact that a small number of vertices results in an overly simplistic network structure, which fails to capture sufficient information from the sequence. Conversely, having too many vertices may lead to information redundancy, which can negatively impact the model's ability to effectively represent the data. 

Overall, parameter variations do not cause significant fluctuations in the performance of the High-TS model across datasets, except for the Wine dataset with very few samples. This demonstrates that High-TS exhibits low sensitivity to parameter changes. Detailed results can be found in Section B.2 of the supplementary materials.

\begin{figure}[] 
\centering
\includegraphics[width=0.48\textwidth]{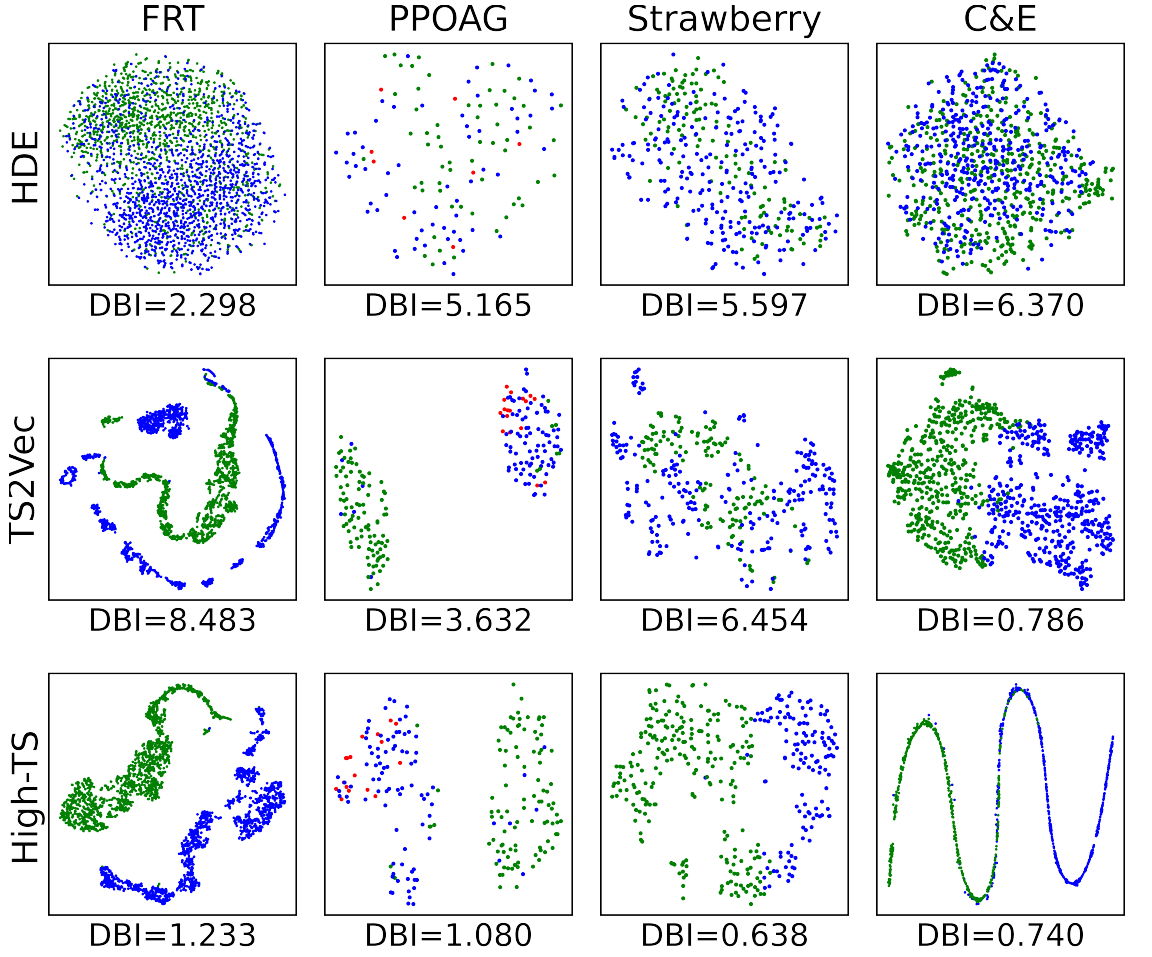} 
\caption{Visualization embedding and Davies-Bouldin Index. Points of different colors belong to different classes within the dataset. Lower DBI values indicate better representation learning performance.}
\label{tsne} 
\end{figure}

\subsection{Visualization}
To provide a more intuitive comparison and further demonstrate the effectiveness of our proposed model, we conduct a visualization task using t-SNE \cite{van2008visualizing} to project the embeddings of the test set into a 2-dimensional space, colored by the true labels. Additionally, we introduce the Davies-Bouldin Index (DBI) \cite{davies1979cluster} to quantify the intra-class similarity and inter-class separability of samples in the 2-dimensional space, where lower values indicate better performance. From the Fig. \ref{tsne}, it is evident that the results of the HDE method are unsatisfactory, as the boundaries between samples with different classes are unclear. TS2Vec performs better than HDE, but in the t-SNE figure of the FRT dataset, green points separate blue points, leading to mixed samples from different classes. High-TS exhibits the best visualization results, where the learned embeddings form the most compact structure with the clearest boundaries between different classes. This proves that High-TS has a stronger ability to learn time series representations compared to other models.

\section{Conclusion}
\label{Conclusion}
We present High-TS, a novel framework designed to capture higher-order interactions in time series through cross-structural embeddings. High-TS utilizes a combination of multiscale Transformer and TDL to model both temporal and spatial perspectives effectively. The integration of contrastive learning further enhances its ability to learn robust and discriminative representations. Experiments demonstrate that High-TS exhibits remarkable generalization ability, maintaining strong performance across different datasets. Future work will focus on refining the scalability of the framework and exploring its applications in other domains.


\bibliographystyle{IEEEtran}
\bibliography{my_reference}

\end{document}